%% file: main.tex
\documentclass[authoryear]{elsarticle}
\usepackage[most]{tcolorbox}
\usepackage{amssymb}
\usepackage[hidelinks]{hyperref}
\usepackage{setspace} 
\usepackage[margin=4cm]{geometry}
\usepackage{enumitem}

\biboptions{authoryear}

\title{A Geolocation-Aware Multimodal Approach for Ecological Prediction}
\author[epfl]{Valerie Zermatten\corref{mycorrespondingauthor}}
\cortext[mycorrespondingauthor]{Corresponding Author: \url{valerie.zermatten[at]epfl.ch}}
\author[epfl]{Chiara Vanalli}
\author[epfl]{Gencer Sumbul}
\author[inria]{Diego Marcos}
\author[epfl]{Devis Tuia}
\address[epfl]{ECEO, Ecole Polytechnique Fédérale de Lausanne (EPFL), 1951 Sion, Switzerland.}
\address[inria]{INRIA, University of Montpellier, Montpellier 34090, France.}

\begin{document}
\begin{frontmatter}
\begin{abstract}
\textbf{1.} Ecological monitoring increasingly relies on heterogeneous data sources, including remote sensing imagery, biodiversity observations, climatic variables, or textual knowledge. While integrating multiple modalities has the potential to improve environmental prediction, current approaches struggle to combine data sources with heterogeneous formats or contents. A central difficulty arises when combining continuous gridded data (e.g., remote sensing) with sparse and irregular point observations such as species records. Existing geostatistical and deep-learning-based approaches typically operate on a single modality or focus on spatially aligned inputs, and thus cannot seamlessly overcome this difficulty.

\textbf{2.} In response to these challenges, we propose a Geolocation-Aware MultiModal Approach (GAMMA), a transformer-based fusion approach designed to integrate heterogeneous ecological data using explicit spatial context. Instead of interpolating observations into a common grid, GAMMA first represents all inputs as location-aware embeddings that preserve spatial relationships between samples. Then, through attention mechanisms, GAMMA dynamically selects relevant neighbours across modalities and spatial scales, enabling the model to jointly exploit continuous remote sensing imagery and sparse geolocated observations.

\textbf{3.} We evaluate GAMMA on the task of predicting 103 environmental variables from the SWECO25 data cube across Switzerland. Inputs combine aerial imagery with biodiversity observations from GBIF and textual habitat descriptions from Wikipedia, provided by the EcoWikiRS dataset. 

\textbf{4.} Experiments show that multimodal fusion consistently improves prediction performance over single-modality baselines and that explicit spatial context further enhances model accuracy. The flexible architecture of GAMMA also allows to analyse the contribution of each modality through controlled ablation experiments. These results demonstrate the potential of location-aware multimodal learning for integrating heterogeneous ecological data and for supporting large-scale environmental mapping tasks and biodiversity monitoring. The source code is included in the submission files.
\end{abstract}
\end{frontmatter}

\begin{figure*}[h]
 \centering
 \includegraphics[width=0.7\linewidth]{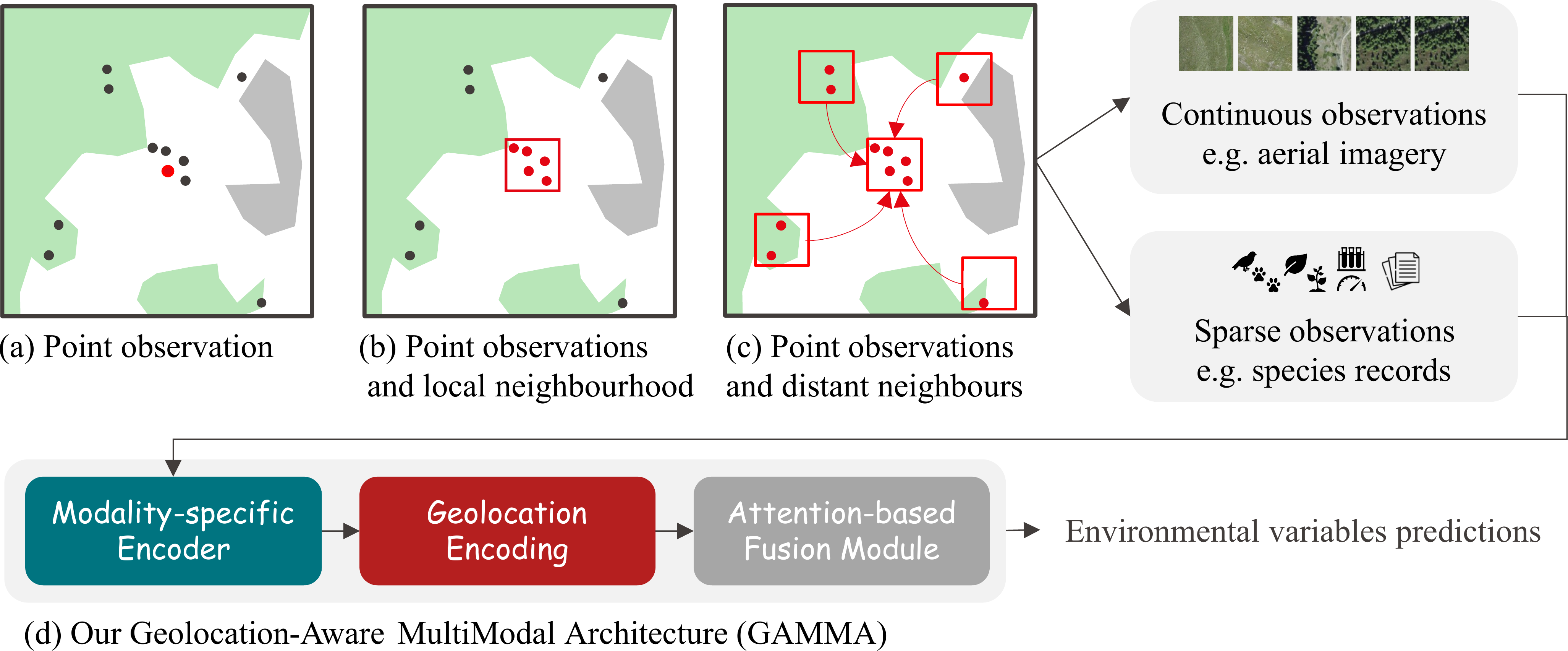}
 \caption{ Overview of the proposed Geolocation-Aware MultiModal Approach (GAMMA). Ecological data assimilation pipelines are often based on single-point observations (a) or on nearby records within a local fixed neighbourhood (b), as commonly exploited in CNNs or GNNs. Leveraging both nearby and distant observations across modalities (c) enables capturing broader spatial dependencies. The proposed GAMMA (d) addresses this limitation by representing observations as location-aware embeddings and using a transformer encoder to dynamically select the most relevant neighbours across modalities and spatial scales. GAMMA effectively combines remote sensing imagery and sparse geolocated text for predicting diverse environmental variables within a single model.
 }
 \label{fig:visual_summary}
\end{figure*}

\section{Introduction}\label{Introduction}

Modern ecological and environmental research increasingly relies on heterogeneous spatial datasets: climatic variables, camera trap photographs, satellite imagery, citizen science records, tabular data, audio, in situ measurements and even text information \citep{cavender-bares_integrating_2022,liu2024perspectives,pollock2025harnessing}. Each modality provides a distinct, but complementary perspective on natural processes, revealing details possibly not available in other sources. Integrating these diverse modalities into a unified framework is widely recognised to increase predictive performance. However, the selection of an appropriate data fusion approach remains a central challenge in data assimilation \citep{isaac2020data,zipkin2021addressing,tuia2022perspectives,miao2026new}. 

A major difficulty in ecological data assimilation arises from their interoperability: the integration of heterogeneous data sources, e.g. continuous and discrete observations, remains a challenge \citep{MARTINEZ2025103485,Jung2023integrated,koldasbayeva2024challenges,wicquart2022workflow}. Continuous sources, such as remote sensing imagery or climatic variables, carry information at every location in a domain and are typically provided as regularly gridded raster data, facilitating spatial alignment across variables. In contrast, many ecological observations are recorded as discrete measurements tied to specific geographic coordinates. Examples include field measurements and species surveys, which are often irregularly distributed in space and describe properties valid over a limited (and often variable) spatial footprint. Combining these heterogeneous spatial representations requires methods capable of handling irregular sampling, exploiting spatial autocorrelation, and propagating information across space.\\

Addressing this challenge has been a long-standing objective in spatial statistics. Geostatistical methods have historically addressed this problem by interpolating sparse observations into continuous spatial fields. Kriging \citep{cressie_origins_1990,matheron1967kriging} or Geographically Weighted Regression (GWR) \citep{fotheringham2009geographically} use spatial autocorrelation to generate predictions at unobserved locations and have long been central tools for environmental (e.g. biomass \citep{wang2005application}) and geological mapping~\citep{SONG2016gwr}, as well as in meteorology (e.g. for snow depth prediction or temperature estimation \citep{cressie_origins_1990}). 

Recent advances in deep learning have revisited these ideas by learning spatial relationships directly from data. Graph Neural Networks (GNNs) \citep{wu2020comprehensive} explicitly model relationships between observations and can learn relevant geographic context by optimally weighting the spatial relationships across training samples. These approaches have been applied to several geospatial tasks, including land use classification \citep{xu_framework_2022} or urban environment analysis \citep{zhu_understanding_2020}. Building on this idea, several studies combine geostatistics with deep learning methods; Kriging Convolutional Networks \citep{appleby_kriging_2020}, for instance, merge the advantages of GNNs and Kriging by directly combining neighbours for spatial prediction tasks, such as bird count modelling or precipitation estimation. Similarly, Deep Kriging Neural Networks \citep{chen_dknn_2024} include a deep learning-based spatial encoder and a Kriging-based decoder for modelling meteorological data. \cite{tian2025Sparse} exploits the Vision Transformer \citep{dosovitskiy2020image} (ViT) architecture to interpolate a digital elevation model from sparse altimetry data. Overall, these methods substantially improve predictive performance over traditional approaches, but remain largely unimodal. 

In parallel to approaches designed for sparse observations, spatial context in regularly gridded data has been extensively exploited through convolutional neural networks (CNNs). CNNs naturally encode local spatial neighbourhoods by applying convolutional filters over raster grids, allowing models to learn spatial patterns directly from neighbouring pixels. This property has made CNNs a dominant approach in remote sensing tasks such as land cover classification \citep{tuia2024artificial}, population mapping \citep{metzger2022fine} or vegetation monitoring \citep{lang2023high}. ViTs have extended this idea by modelling spatial context through self-attention over image patches, enabling the integration of larger spatial contexts while preserving the grid structure of remote sensing imagery \citep{lu2025vision,xiao2025foundation}. Overall, these approaches typically address either dense raster data or sparse point observations, but not both simultaneously.\\

Instead of interpolating geospatial information from observations, an alternative line of work incorporates spatial information directly into the model through location encoding. Location encoding methods transform geographic coordinates into high-dimensional embeddings that capture spatial relationships between locations \citep{mac2019presence,mai_review_2022,cole2023spatial,russwurmgeographic}. 
Beyond geographic proximity, these embeddings can also incorporate contextual information derived from auxiliary data such as satellite imagery \citep{klemmer_satclip_2024}, ground-level imagery \citep{mai2023csp} or climatic variables \citep{daroya2025wildsat}. Location embeddings are beneficial in many geospatial applications, from species distribution modelling, fine-grained species recognition, to climate variables estimation \citep{cole2023spatial,chu2019geo,lange_feedforward_2025}. 

\enlargethispage{\baselineskip}
Transformers \citep{vaswani2017attention}, which excel at integrating heterogeneous sequences of inputs, offer a natural solution to integrate multiple data modalities. Its architecture has proven highly effective in vision-language models such as VisualBERT \citep{visualBERT_li2020does} or LLaVA \citep{liu2023llava}, which process image and text representations jointly, enabling rich cross-modal reasoning. This same principle can be applied to geospatial data, where heterogeneous inputs can be treated as complementary representations describing a shared context. 

Two recent approaches leveraging transformers for geospatial prediction are particularly related to our work. Space2Vec \citep{maimulti} uses a transformer architecture to incorporate location embeddings of multiple geographic objects to predict  Points of Interest (POIs) type or image category based on their geolocation. However, similarly to Kriging, Space2Vec uses neighbouring target values as inputs (e.g. POI types of neighbouring locations), restricting its applicability to settings where such labels are not available.
\cite{lange_feedforward_2025} presents an attention-based approach to predict species range maps based on multiple geolocated observations in a few-shot setting. This approach is primarily coordinate-centric, without exploring relative positional encodings, and adding, optionally, one non-geocoded sentence or image as metadata.  Addressing these limitations calls for a framework that requires no neighbour supervision, is inherently multimodal, and explicitly encodes spatial relationships across heterogeneous input sources.

We propose a \textbf{G}eolocation-\textbf{A}ware \textbf{M}ulti\textbf{M}odal \textbf{A}pproach (GAMMA)  that integrates multimodal ecological information sources using explicit location encoding. As illustrated in Fig. \ref{fig:visual_summary}, GAMMA learns spatial context beyond a single location (\ref{fig:visual_summary}(a)) or the fixed local neighbourhood typically captured by CNNs (\ref{fig:visual_summary}(b)). Instead, it leverages sparse observations and gridded data within an arbitrarily large neighbourhood (\ref{fig:visual_summary}(c)). This is implemented as location embeddings that preserve the spatial arrangement between samples. Rather than interpolating discrete sources into a continuous spatial field, GAMMA represents all inputs as location-aware representations and uses a transformer encoder to learn cross-modal spatial relationships directly from data. Similar to a graph-based approach or to GWR, GAMMA dynamically selects the most relevant neighbours by including a notion of geographic distance, and fuses them according to feature similarity. Furthermore, GAMMA naturally accommodates flexible input combinations, enabling systematic ablation of individual modalities to quantify the independent contribution of each data source.

We evaluate  GAMMA on the prediction of 103 environmental variables from the SWECO25 \citep{kulling2024sweco25} data cube. SWECO25 variables span hydrological, climatic, geological, vegetation, population, land cover, and soil properties, and together form a comprehensive characterisation of a location that no single sensor can capture alone. As input modalities, we integrate remote sensing imagery as gridded variables with biodiversity observations from GBIF and textual habitat descriptions from Wikipedia, sourced from the EcoWikiRS \citep{zermatten2025ecowikirs} dataset. Similar to recent works \citep{daroya2025wildsat,lange_feedforward_2025,hamilton2024lesinr}, the species occurrence records serve as a means to assign a geolocation to Wikipedia-derived habitat sentences. Text provides insight into the local environmental conditions, which are potentially unavailable from the remote sensing perspective. SWECO variables are extracted at the locations corresponding to EcoWikiRS samples, forming a benchmark dataset that we release to address the current lack of benchmarks for vision-language models in geospatial regression tasks \citep{xue2025regression}.

The flexible input structure of GAMMA enables us to quantify the independent contribution of each modality, which is rarely systematically studied in vision and language models for ecological applications \citep{daroya2025wildsat,hamilton2024lesinr,lange_feedforward_2025,miao2026new, xue2025regression,gu_bioclip_2025,stevens_bioclip_2024,wen_phenology_2025}. Here, we evaluate text-only, image-only, and combined settings, isolating what ecological knowledge each source provides. Our experiments demonstrate that multimodal fusion outperforms single-modality baselines, and explicit spatial context encoding yields additional synergistic gains over location-agnostic approaches. The proposed approach has the potential to address challenges related to the incorporation of sparse records with gridded data.

While focusing on vision and language inputs, GAMMA is broadly transferable to other domains and types of geospatial inputs in ecology or beyond. To summarise, this work makes three contributions:

\begin{itemize}[itemsep=0pt, topsep=0pt, parsep=0pt, partopsep=0pt]
 \item We propose GAMMA, a geolocation-aware multimodal approach for integrating sparse and continuous spatial data.
 \item We release an open benchmark dataset of 103 environmental variables combining EcoWikiRS and SWECO25 for spatial regression.
 \item We provide a systematic evaluation of the contribution of textual ecological knowledge relative to remote sensing imagery.
\end{itemize}

\section{Materials and Methods}\label{sec:method}

\begin{figure}[t]
 \centering
 \includegraphics[width=0.8\linewidth]{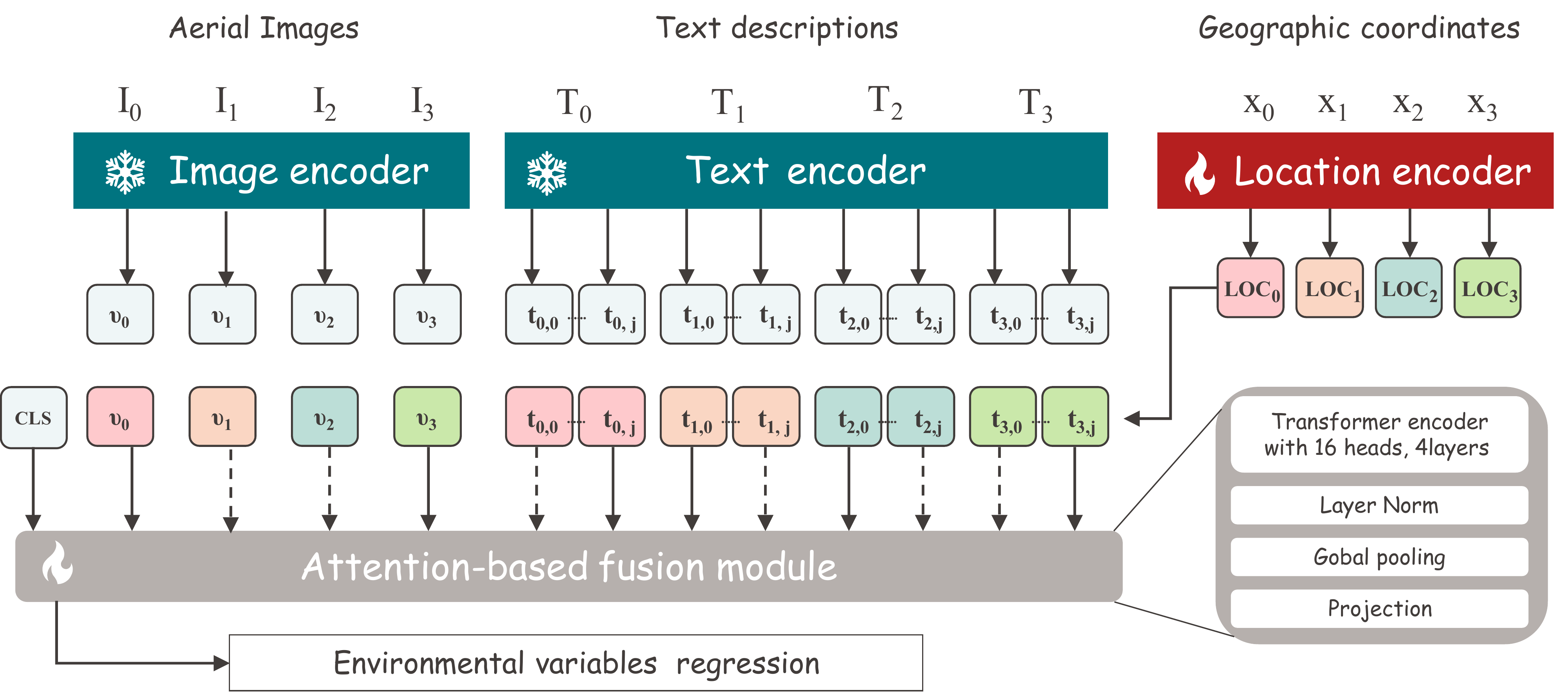}
 \caption{The proposed model architecture is illustrated with a spatial context of 3 nearest neighbours (k=3), with images, text and their respective geolocations as inputs. The pretrained encoders for image and text are frozen, while the location encoder and the transformer-based fusion module are trained. Random token masking (dashed arrows) is applied on the generated visual ($v_i$) and textual ($t_{i,j}$) embeddings after the fusion with the location encoding ($LOC_i$). The final predictions are derived from the classification token (CLS) at the output of the transformer module. Colours for $LOC_i$ tokens denote distinct geographic locations.}
 \label{fig:architecture}
\end{figure}

Through the predictions of 103 SWECO environmental variables, GAMMA solves a multiple regression task by integrating sparse geolocated text and aerial images. The overall model architecture is shown in Fig.~\ref{fig:architecture}. We adopted a modular architecture that consists of an image encoder, a text encoder, a location encoder, and a transformer-based fusion module. 

\subsection{Preliminaries} 
We consider a dataset composed of $N$ geolocated samples with the following elements: the location $x_i \in \mathbb{R}^2$ is a GPS coordinate; $I_i \in \mathbb{R}^{3 \times h \times w} $ is a remote sensing image of location $x_i$ with RGB bands; $T_{i}=\{s_{i,0},...,s_{i,j}\}$ is the set of $j$ text sentences associated with location $x_i$; and $y_i \in \mathbb{R}^m$ is the regression target, while $m$ denotes the number of environmental variables to be predicted. 
 
For each location $x_i$, we compile the list of $k \in \mathbb{N}$ nearest samples (or $k$ nearest neighbours, kNN). The neighbours are defined in terms of Euclidean distance between the respective geographic coordinates, sorted by increasing distance, while $k=0$ denotes the index of the location of interest. Thus, a training sample $p_i$ is made of $k+1$ geographic coordinates, i.e. one for the location of interest and $k$ for its NN, and, following the same logic, $k+1$ remote sensing images and $k+1$ sets of sentences. Thus, each training sample integrating context from $k$ neighbours can be written as follows: 

$$p_{i}= \{x_{i,0},...,x_{i,k}, I_{i,0},...,I_{i,k}, T_{i,0},...,T_{i,k} \}$$

The proposed approach aims at learning a model $f$ that estimates the environmental variables $y_*$ for a new location $x_*$, as $ \hat{y}_* = f (p_*)$.

\subsection{Text and image encoding} \label{sec:method_archi}
 
Given a training location $p_i$, a pretrained visual encoder $f_v$ encodes the $k+1$ remote sensing images into $k+1$ visual embeddings $(v_i)_{i=0}^k$ with $v_i \in \mathbb{R}^{d}$. For each location, a number $j$ of sentences is then selected. The selected sentences are fed to a tokeniser and to a pretrained text encoder $f_t$, producing $j (k+1)$ textual embeddings $(t_i)_{i=0}^{j(k+1)}$ with $t_i \in \mathbb{R} ^d $. We found empirically that selecting the top-$j$ sentences per location by computing the cosine similarity with the visual embeddings using a pretrained VLM is the most effective way to integrate relevant text information. More details are available in Section \ref{sec:ablation}.

\subsection{Location encoding} 
\label{sec:method_location}

We encode the spatial structure of each textual and visual embedding by combining it with a geolocation encoding based on their geographic coordinates ($LOC_i$ in Fig. \ref{fig:architecture}). While several studies propose pretrained location embeddings available at a global scale, we consider them unsuitable for our study area, which is limited to a single country (Switzerland). Such embeddings would likely produce overly homogeneous representations. Following \cite{maimulti}, we compare several learnable or fixed location encoding approaches (see Fig. \ref{fig:positional_encoding}), while the results are discussed in Section \ref{sec:ablation_location}:

\begin{figure}[t]
 \centering
 \includegraphics[width=0.8\linewidth]{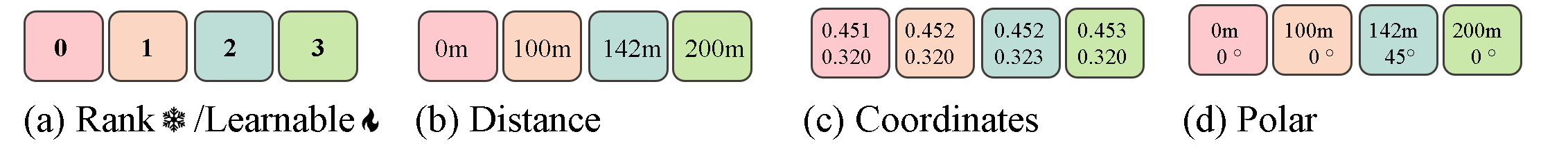}
 \caption{Visualisation of what numerical values are encoded through location embeddings using
three spatial neighbours (k=3)}
 \label{fig:positional_encoding}
\end{figure}

\begin{itemize}
 \setlength\itemsep{0em}
 \item \textbf{None}: No geolocation information is provided.
 \item \textbf{Rank}: We encode the rank of nearest neighbours, i.e. the k in the kNN, with a $512$ dimensional $1-D$ sinusoidal encoding,
 similarly to the Transformer 1-D positional encoding \citep{vaswani2017attention}. Embeddings are concatenated with the text or visual features and merged through a fully connected (FC) layer.
 \item \textbf{Learnable}: We encode the rank of nearest neighbours for both text and image embeddings with learnable embeddings, which are then summed with the corresponding vision or text features. The embeddings are initialized with a uniform distribution between $ [ -1/h_{dim}^{1/2}, 1/{h_{dim}^{1/2}} ]$.
 \item \textbf{Distance}: We encode the Euclidean distance in meters from the origin to the target nearest neighbour with $1-D$ sinusoidal encoding with $h_{dim}=512$ dimensions up to a maximum distance of $10,000~m$. We concatenate the distance embeddings with their corresponding text or visual embeddings and combine them through a FC layer.
 \item \textbf{Coordinates}: We encode the geographic coordinates for each location with a $2-D$ sinusoidal multi-scale encoding, similarly to the ``grid'' approach in \cite{maimulti}. The geographic coordinates are scaled to the range $[0,1]$ based on the min-max coordinates of the study area. We encode each coordinate, i.e. North and East, separately with $h_{dim} /2 =256$ dimensions. We concatenate them with their corresponding text or visual embeddings, and finally merge them through a FC layer.
 \item \textbf{Polar}: We combine the Euclidean distance from the origin to the target nearest neighbour with the angle between the positive x-axis and the vector pointing from the origin to the nearest neighbour. We use sinusoidal functions to encode both the distance and the angle, similarly to the ``polar'' approach from \cite{maimulti}. We concatenate the distance, the angle and the features embedding together and combine them through a FC layer.
 
\end{itemize}

After encoding the location for all inputs, each text and visual embedding can be considered as an input token without specific order, since the position relative to the origin is preserved.

\begin{figure}[t]
 \centering
 \includegraphics[width=0.9\linewidth]{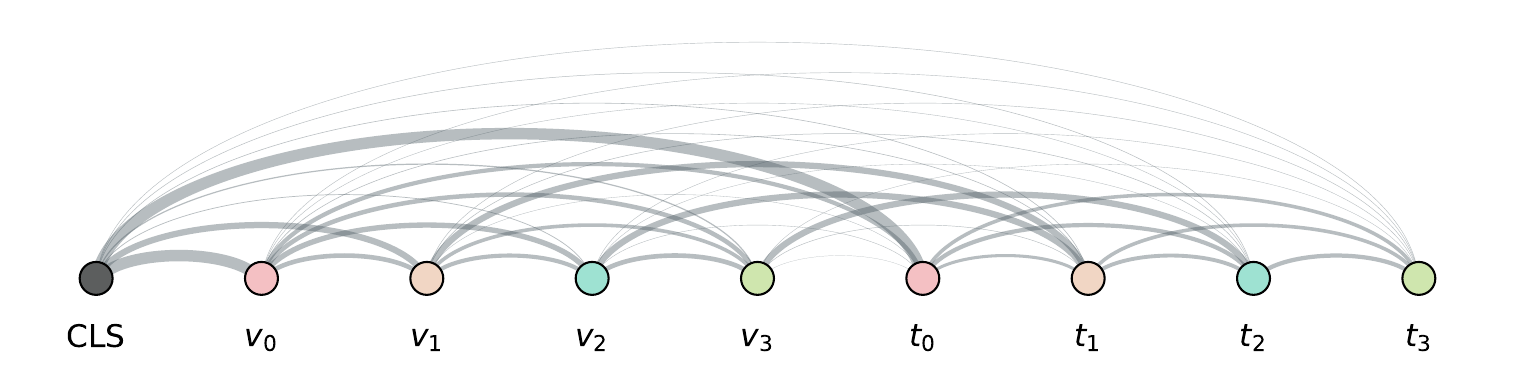}
 \caption{Illustration of the attention between multimodal tokens in the fusion module, where nodes represent the global classification token (CLS), image tokens $v_i$ and text tokens $t_i$, and edge width denotes pairwise attention strength. }
 \label{fig:arc_diagram}
\end{figure}

\subsection{Multi-modal fusion with the transformer-based module}
Once all inputs have been encoded into a common embedding space, fusion can be performed. The fusion module is composed of a transformer encoder, a layer-normalisation layer, a global pooling layer on the global classification (CLS) token and a projection layer that leads to the final model predictions. The attention mechanisms in the transformer encoder enable the fusion of the multi-modal tokens. The motivation behind this design is that the attention mechanism dynamically weights spatial neighbours (see Fig. \ref{fig:arc_diagram}), emphasising the most informative neighbours, while ignoring distant or irrelevant ones, thereby learning a context-aware spatial representation.

\subsection{Dataset}\label{sec:settings} 
This section describes the datasets used in our experiments. 

\subsubsection{EcoWikiRS}\label{sec:dataset} 
The EcoWikiRS \citep{zermatten2025ecowikirs} dataset contains over $90,000$ text-image pairs, combining high-resolution RGB aerial imagery from Switzerland ($50~cm$ resolution, $100~m\times100~m$) with sentences from Wikipedia describing the habitat of a species observed at the location of the images. The geolocation of species observations from the Global Biodiversity Information Facility (GBIF)\footnote{\url{https://www.gbif.org/}} was used to pair species habitat descriptions with aerial images. A specificity of this dataset is that images are associated with more than one sentence, since several sentences are usually extracted from each Wikipedia article. 
The dataset is split into train, validation and test following a spatial block split, with a block size of roughly $20~km$, and nearest neighbours are sampled only within the same split. This avoids performance inflation due to spatially autocorrelated training and test samples \citep{kattenborn_spatially_2022}.

\subsubsection{SWECO25} 
We assess the contribution of each modality in the EcoWikiRS dataset by evaluating its capacity to predict environmental variables. For this task, we use the SWECO25 \citep{kulling2024sweco25} dataset, a data cube of more than $5,000$ environmental rasters at $25~m$ spatial resolution covering Switzerland. For each location in EcoWikiRS, we extract average values covering the surface of the input image ($100\times100~m$) for $103$ selected SWECO25 variables. We standardise the numerical values across the training set with zero mean and unit variance. The $103$ variables are grouped into seven thematic groups, which are described below, while the Supplementary Section \ref{sec:sweco_docu} provides more details on the selection, preprocessing and individual description of the SWECO25 variables:

\begin{itemize}
 \setlength\itemsep{0em}
 \item Bioclimatic: Bioclimatic variables from WordClim.
 \item Geologic: Subsoil classified according to lithological criteria.
 \item Hydrological: Distance to the hydrological network and ecomorphological state of rivers.
 \item Edaphic: Ecological indicator values for soil properties and nitrogen and phosphorus loads.
 \item Land use and cover: Land-use and cover classification. 
 \item Population: Human population density and distance to transportation network. 
 \item Vegetation: Vegetation height and dominant leaf type.
\end{itemize}

\subsection{Experimental setup}

The model is trained for $100$ epochs to minimise a mean squared error (MSE) loss, a standard loss for regression, by using the AdamW optimiser \citep{loshchilovAdamW} with the initial learning rate of $1e^{-4}$ and the mini-batch size of $256$. 
We use the pretrained text and vision encoders from SkyCLIP-ViT-B \citep{wang2024skyscript}, following its effectiveness on the EcoWikiRS dataset \citep{zermatten2025ecowikirs}. We also explored the use of a general vision encoder for encoding the image and a pretrained LLM to produce text representations, but found empirically that the alignment learned with contrastive learning between text and image features, as in SkyCLIP, provides a more accurate multi-modal fusion in our approach. We empirically observed that fine-tuning or not the visual and textual encoders does not lead to significant performance differences; thus, we keep both of them frozen. 

During training, random token masking is applied on the visual and textual tokens, with the probability of $30\%$, to reduce overfitting and increase model robustness to varying NN distances. We use the \textit{habitat} sentences from the EcoWikiRS data, i.e. sentences relative to the habitat section of the article only, since they were shown to perform best in \cite{zermatten2025ecowikirs}. We use the four sentences with the highest cosine similarity to the image in the SkyCLIP latent space, i.e., the top-j sentence with $j=4$. If fewer than four sentences are available, sentences are chosen with replacement. If not mentioned otherwise, a \textit{polar} location encoding is applied. We evaluated the regression of environmental variables with the coefficient of determination ($R^2$).

\section{Results}\label{sec:results}

The performance of the proposed approach is assessed against single-location and single-modality baselines in Section \ref{sec:POC}. Then, an analysis by thematic group of environmental variables is presented in Section \ref{sec:thematic}, which highlights the different contributions of text and images for specific environmental variables. Finally, Section \ref{sec:ablation} analyses the contributions of the various model components to the overall performance, such as text inputs (number of sentences, text selection) and location encoding approaches.

\subsection{Combining text and images in a spatial context} 
\label{sec:POC}

\begin{figure}[t]
 \centering
 \includegraphics[width=\linewidth]{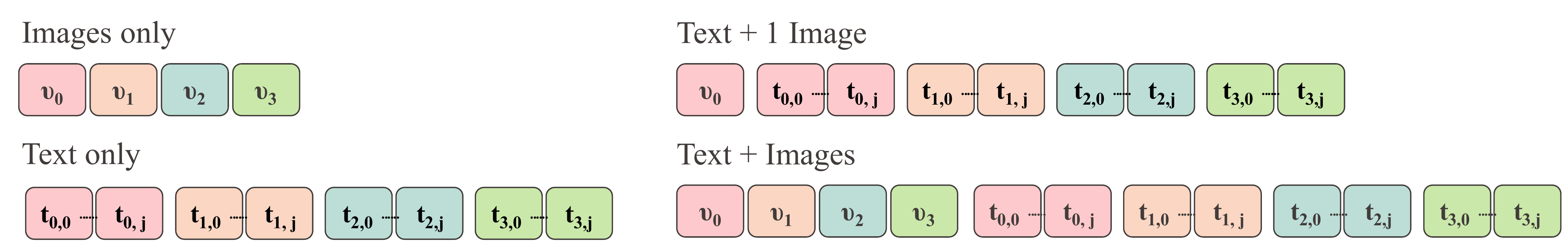}
 \caption{Different experimental settings used to explore various modality combinations, using three spatial neighbours (k=$3$) and four sentences per location ($j=4$). Colours denote distinct geographic locations.}
 \label{fig:all_settings}
\end{figure}

We exploit the architectural flexibility of our model for systematically testing how different modalities and different numbers of neighbours affect the predictive performance. Specifically, we compare several settings using data from a single location (without spatial context) versus several nearest neighbours (with spatial context) and evaluate the effect of combining text and images. All settings are shown in Fig. \ref{fig:all_settings}. The ``Text only'' and the ``Images only'' settings use exclusively one modality, whereas both modalities are mixed together in the ``Text + Images'' experiment. We also observe the contribution of a single image location, with text (``Text + 1 Image''), since adding a single image at the location of interest provides a significant contribution in many cases.

\begin{figure}[t]
 \centering
 \includegraphics[width=0.7\linewidth]{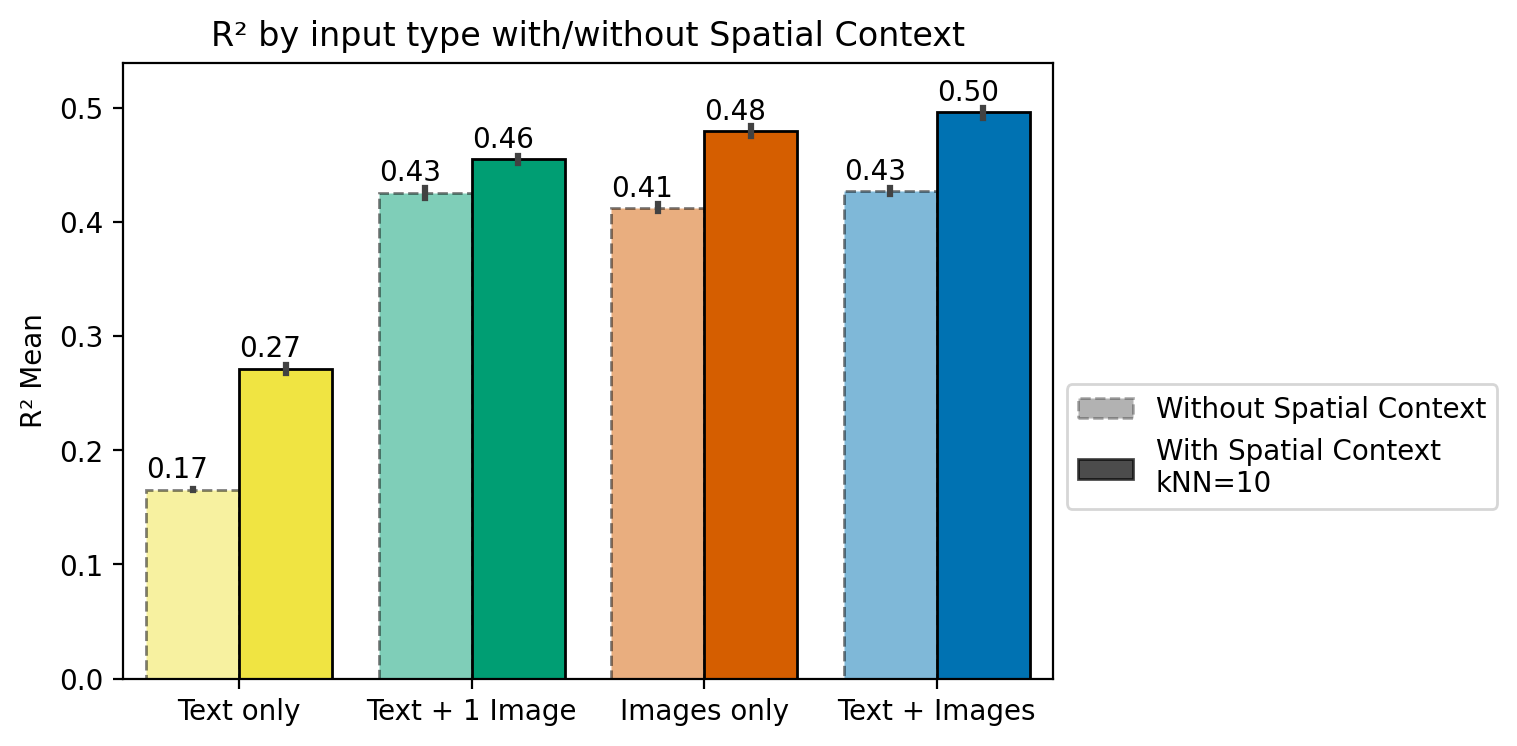}
 \caption{Mean coefficient of correlation $R^2$ on the EcoWikiRS test set for predicting SWECO25 variables, with images, text or their combination as input, comparing with or without spatial context ($k=10$). The black line over each bar indicates the standard deviation across five random seeds.}
 \label{fig:w_wo_spatial_context}
\end{figure}

Fig.~\ref{fig:w_wo_spatial_context} shows the mean coefficient of correlation $R^2$ on the EcoWikiRS test set for predicting the environmental variables with the standard deviation over five random seeds. The best performances are obtained when combining text and images, both with spatial context, with an $R^2$ score of $0.50$ (``Text + Images'') or without spatial context with an $R^2$ score of $0.43$ (``Text + 1 Image'', ``Text + Images'', i.e. equivalent setting without NN). This confirms that incorporating text with images brings complementary information, and combining them is key to reaching the highest performance. 
The comparison between ``Text only'' and ``Image only'' models demonstrates that images carry richer information for the considered tasks, both with or without spatial context. This is confirmed with the ``Text + 1 Image'' setting, which shows that the corresponding RS image at origin carries an important share of the information, and offers a significant boost in performance compared to the ``Text only'' setting. Integrating spatial context ($k=10 $) consistently improves predictive performance across all settings, compared to using data from a single location. The ``Text only'' approach benefits the most from the integration of context, with its performance rising from $0.17$ to $0.27$ ($+58\%$). \\

\begin{figure}[t]
 \centering
 \includegraphics[width=0.7\linewidth]{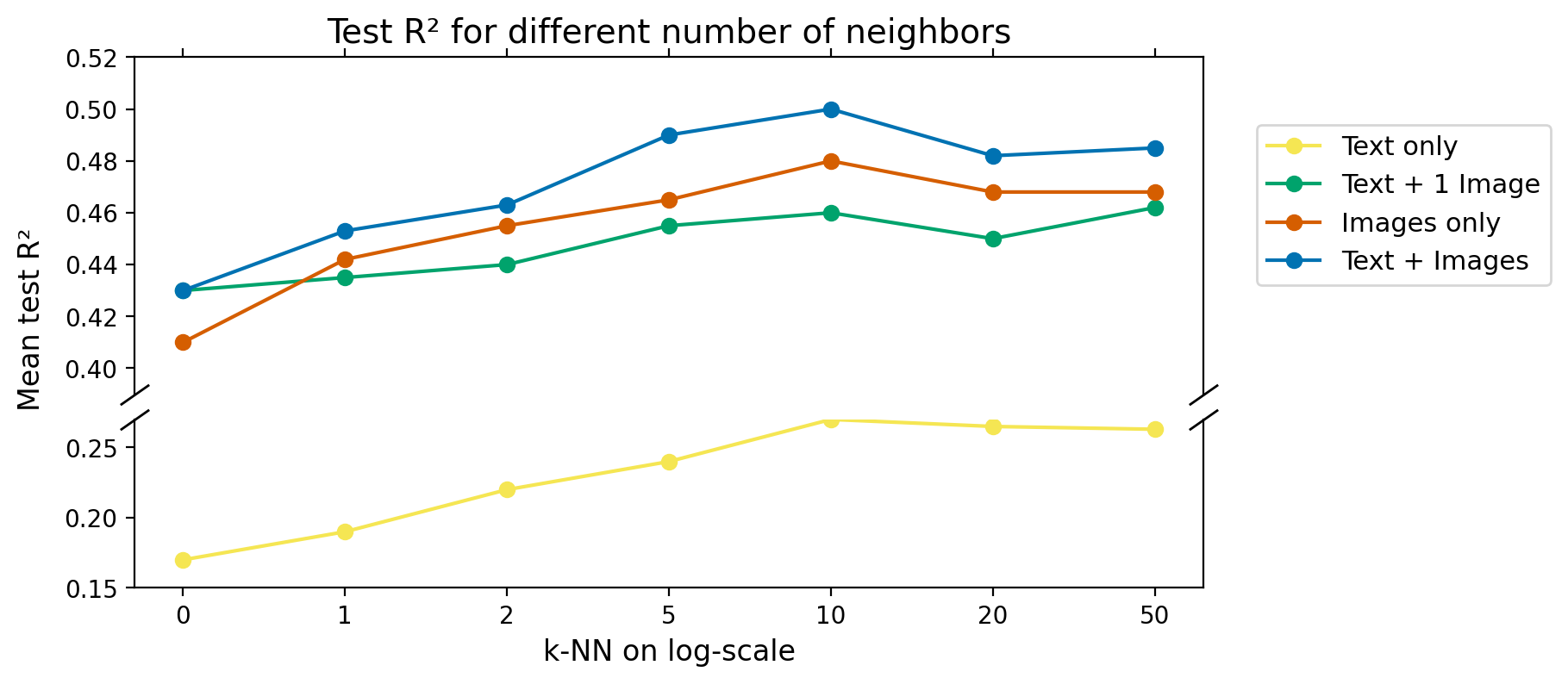}
 \caption{Mean coefficient of correlation $R^2$ for predicting SWECO25 variables on the EcoWikiRS test set, when adding incrementally more neighbours. The axis is discontinuous between $0.27$ and $0.40$ for clarity of presentation.}
 \label{fig:number_of_NN}
\end{figure}

\begin{figure}[t]
 \hspace{2.2cm}
 \includegraphics[width=0.5\linewidth]{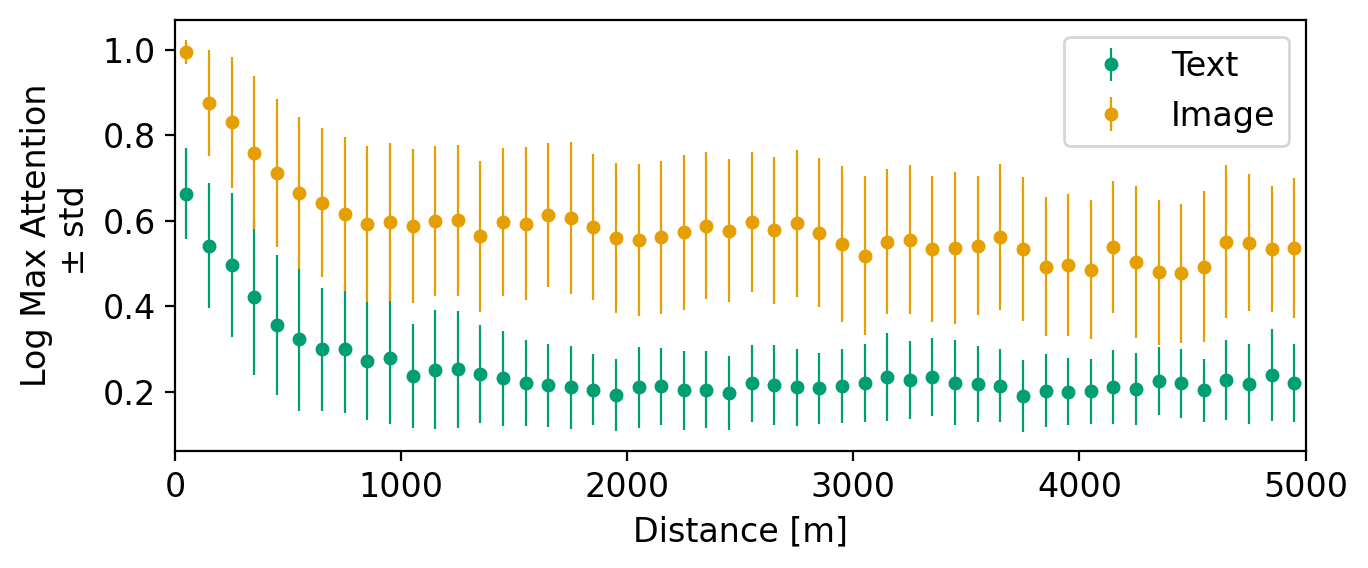}
 \caption{Attention visualisations for the ``Text + Images'' model on the EcoWikiRS test set. Attention values are log-maximum scores taken across all layers and scaled to $[0,1]$, and they are represented as means and standard deviations, per distance bins of $100~m$.}
 \label{fig:attention_dist} 
\end{figure}

Fig.~\ref{fig:number_of_NN} illustrates how the number of nearest neighbours affects the accuracy metrics. Performance increases steadily up to $10$ NN, after which it stagnates or even drops for some models. 
Including distant neighbours yields no meaningful performance gain while increasing computational cost, arguing for a conservative approach to data integration.
This behaviour is becomes clearer when looking at the Suppl. Figs ~\ref{fig:mean_dist_to_knn2}-\ref{fig:similarity_vs_distance}, which illustrates the mean distance to NN, and the spatial autocorrelation between input observations, measured as the feature similarity as a function of distance.
Higher similarity between features occurs up to a distance of $1,000~m$ for visual embeddings, and $700~m$ for the text ones, on average, which is close to the average distance of the $10^{th}$ NN, which equals $880~m$. This suggests that the spatial attention mechanism implicitly captures spatial dependencies and effectively uses them to weight input features.

Fig.~\ref{fig:attention_dist} confirms this hypothesis and shows the attention for all visual and text tokens as a function of distance, in the test set. The attention score is computed as the logarithm of the maximum attention score across all transformer encoder layers for the best ``Text + Images'' model and plotted as distance bins of $100~m$. Both modalities exhibit their highest attention values at short distances, which then monotonically decrease up to a distance of about $1,000~m$. The visual tokens exhibit, overall, a higher attention score compared to the text tokens.

Overall, this supports a conservative yet clever integration of features, while leveraging a few NNs and a single image already achieves a notable gain in performance compared to methods involving a single modality, or without spatial context.

\subsection{Specific contribution of text and images for different types of environmental variables} \label{sec:thematic}

\begin{figure}[t]
 \centering
 \includegraphics[width=0.85\linewidth]{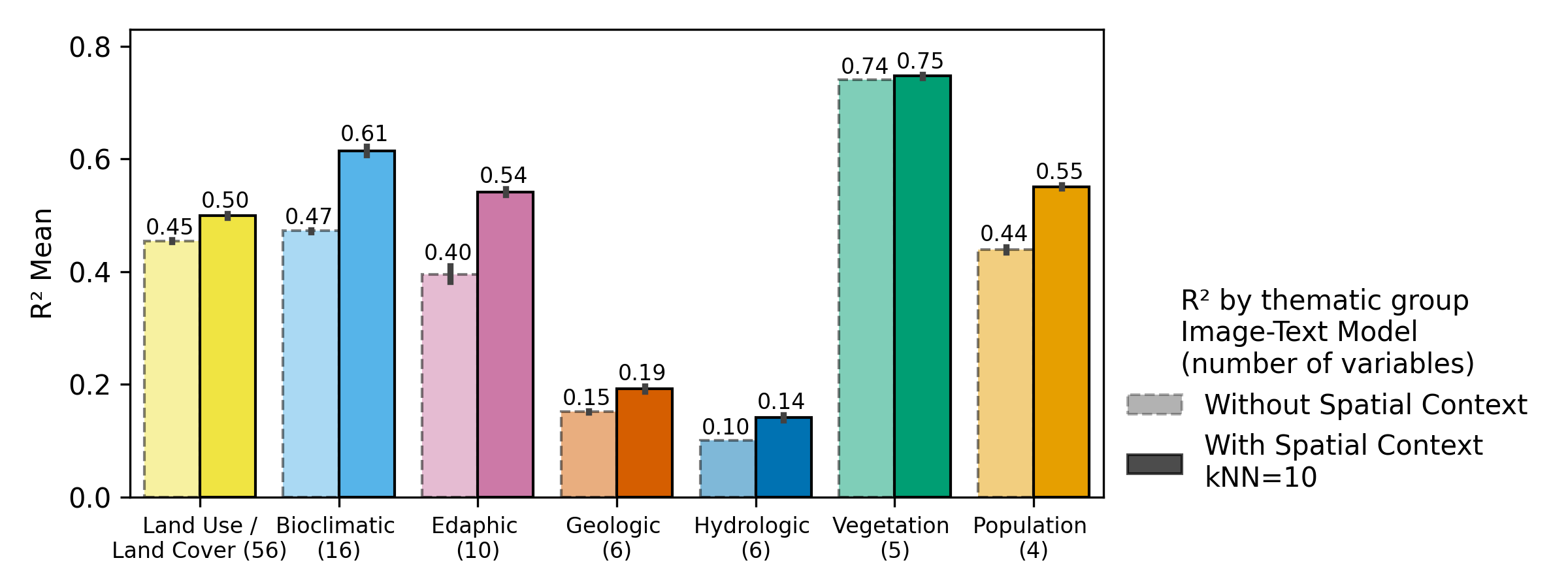}
 \caption{Comparison of model performance on different groups of environmental variables when incorporating spatial context with $10$ neighbours. The black line over each bar indicates the standard deviation across five random seeds.}
 \label{fig:thematic_contribution}
\end{figure}

\begin{figure}[h!]
 \centering
 \includegraphics[width=0.7\linewidth]{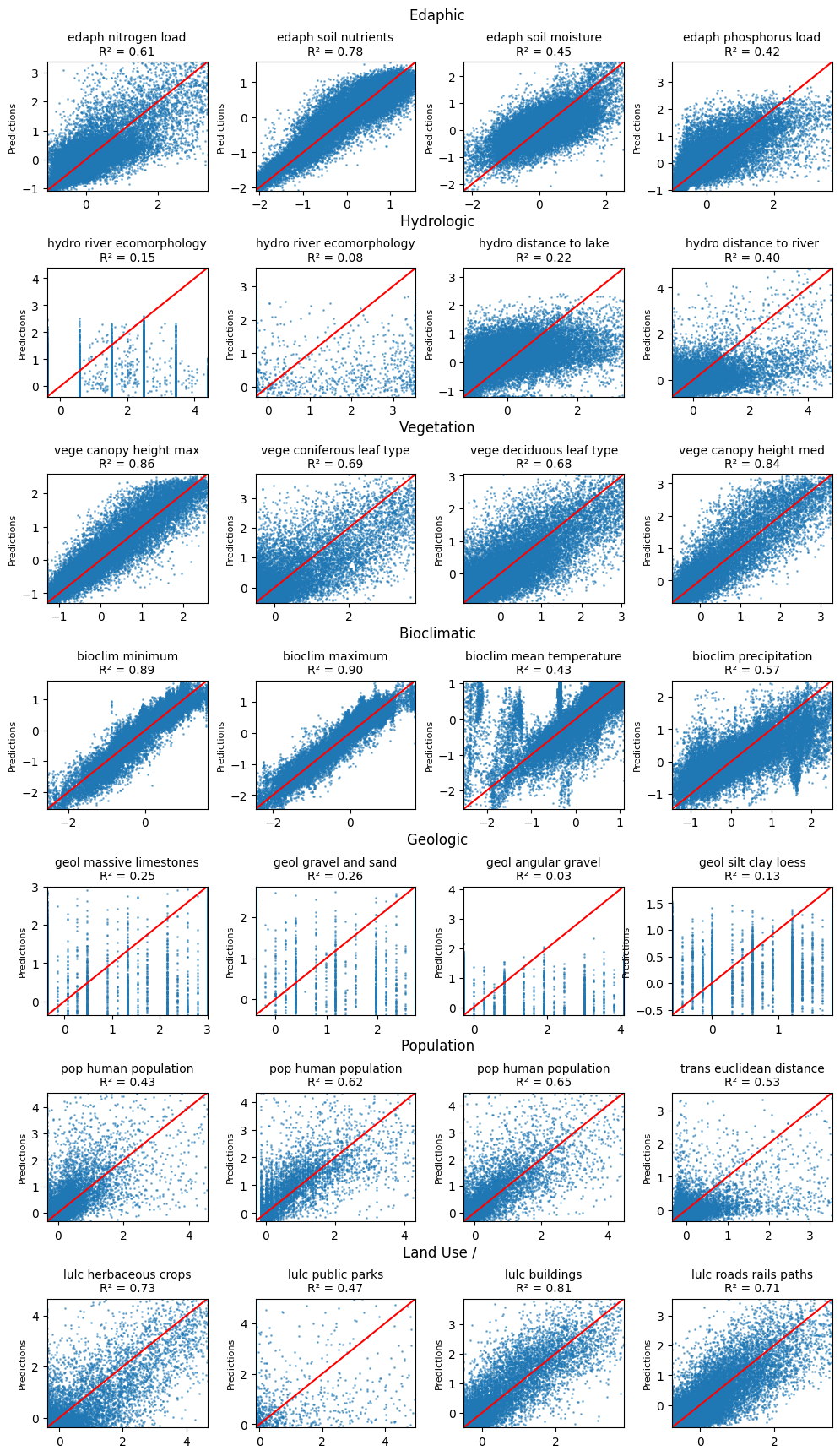}
 \caption{Scatter plots of SWECO25variables and their predicted values. Four variables per thematic group are randomly plotted. The red line represents the $x=y$ line. Predicted values between 1\% and 99\% percentiles are plotted (excluding outliers for readability reasons).}
 \label{fig:scatter_sweco}
\end{figure}

Fig.~\ref{fig:thematic_contribution} shows the comparison of the regression performances of our best model, i.e. ``Text + Images'', with or without spatial context, for the SWECO25 variables grouped into seven thematic groups. Additional scatter plots of predicted values against SWECO25 values are presented in the Suppl. Fig. \ref{fig:scatter_sweco} to give more insight into performance at the variable level. 

For all the thematic groups, a gain in performance is obtained when adding spatial context, but to a varying degree. The land use/land cover, bioclimatic, edaphic and population groups are associated with globally high $R^2$ scores, and received a significant boost when adding spatial context, with a large relative increase in performance from $0.47$ to $0.61$ ($+30\%$) for bioclimatic variables and from $0.40$ to $0.54$ ($+35\%$) for the edaphic group. The overall highest $R^2$ scores are obtained for the Vegetation group, both with ($0.75$) and without ($0.74$) spatial context, while this small relative improvement when adding spatial context suggests that features at the origin are sufficient to model this group. 
\enlargethispage{\baselineskip}
For the remaining groups, \textit{geological} and \textit{hydrological} variables, the model performance remains low in both scenarios, with only a slight increase with spatial context. These factors are discretised from categorical variables and seem hardly predictable from the RS perspective; adding context information only provides a moderate benefit.\\

\begin{figure}[t]
\centering
\includegraphics[height=5cm]{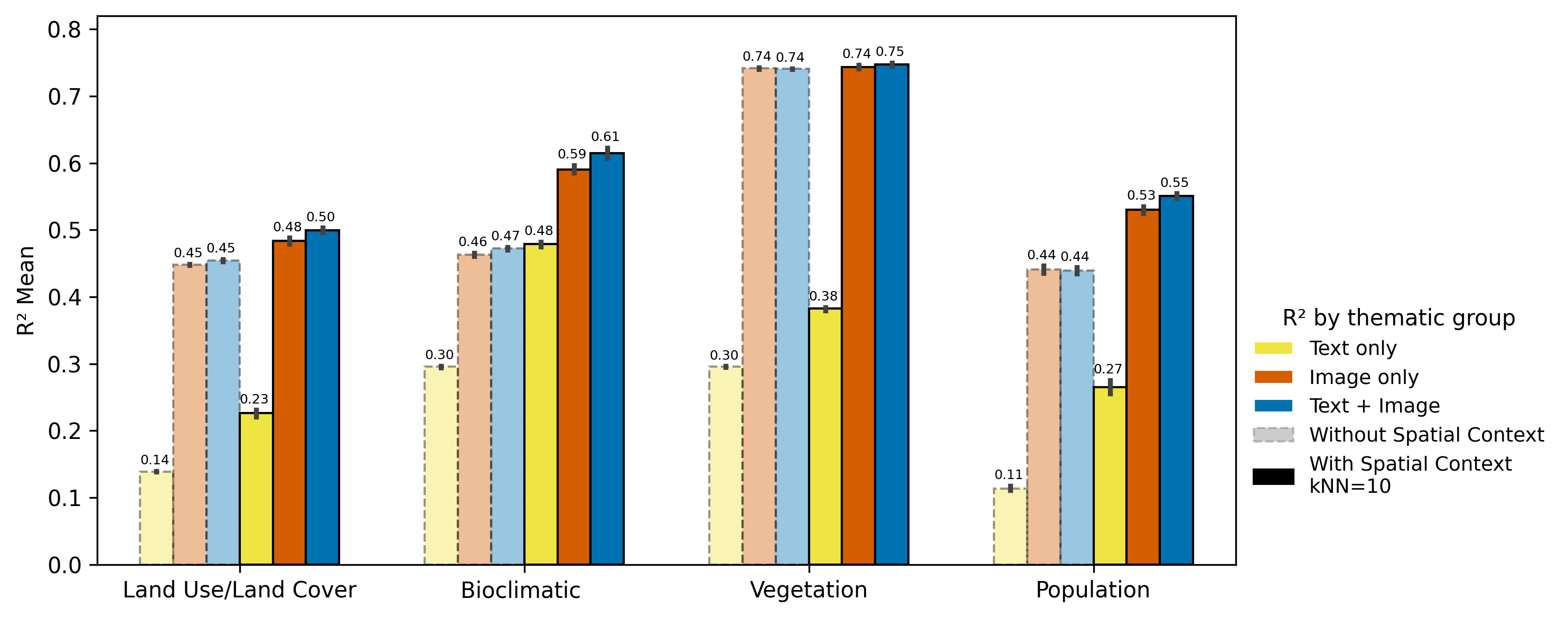}
\includegraphics[height=5cm]{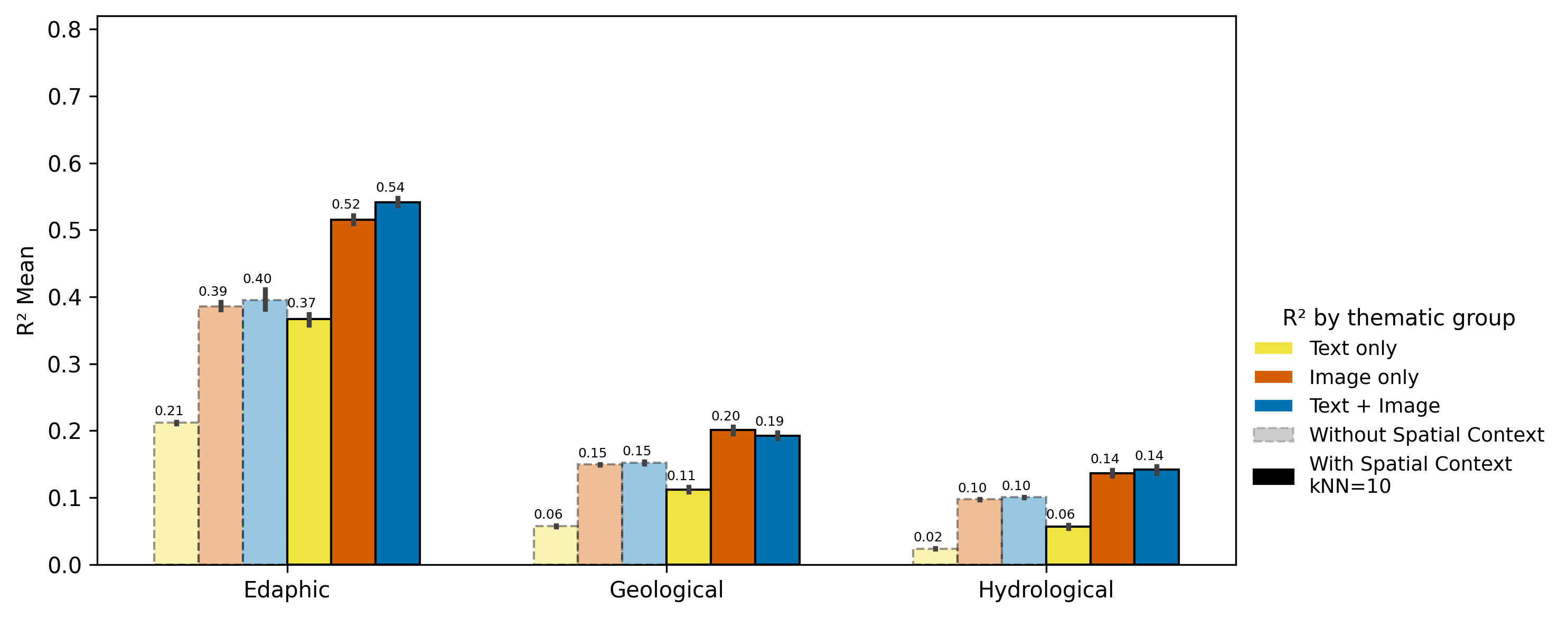}
\caption{Comparison of performance for the groups of environmental variables when incorporating different modalities with or without spatial context. The black line over each bar indicates the standard deviation across five random seeds.}
\label{fig:thematic_contri_interest}
\end{figure}

Fig.~\ref{fig:thematic_contri_interest} compares the different settings per group of environmental variables. Overall, the ``Text + Images'' setting with spatial context performs the best, closely followed by the ``Image only'' setting, although the gap varies across groups. This confirms a complementary effect from combining modalities, meaning that fusing information from textual and visual data provides a more robust predictive model than either modality alone. 
The overall weakest performance is obtained again for the geological and hydrological groups, showing a neutral or even negative effect when adding text, which we interpret as the absence of sufficient information for modelling them in the text data. 

For vegetation, we observe that using a single image (``Image only'', without spatial context) performs remarkably well, with almost no improvement from either adding text or context, which indicates that these variables are essentially captured by the visual features at the origin. 

Clear improvement from ``Text + Images'' over the next best single modality (``Image only'') is observed for land use/land cover, bioclimatic, population and edaphic variables, suggesting the textual data is most beneficial in these domains. Characteristics describing these variables are more consistently mentioned in textual descriptions of species' habitat preferences. The presence of useful information is also shown with the relatively strong performance of the ``Text only'' model for the edaphic and bioclimatic groups. However, the population and land use/ land cover variables benefit from incorporating text data only when combined with spatial context.

\subsection{Ablation studies}\label{sec:ablation}
The ablation studies compare several methods to integrate text and location information in the proposed pipeline.

\subsubsection{Integrating text}\label{sec:ablation_text}
We investigate how to best integrate information from multiple text snippets in our pipeline. We compare three strategies for selecting textual data. First, we increase the number of sentences per location by randomly choosing $j \in [0,1,2,4,8,16 ]$ sentences from the location of interest. As the second step, instead of using randomly selected sentences, we choose the top-$j$ with $j \in [0,1,2,4 ]$ sentences at the location of interest based on the text-image cosine similarity. This ensures that the textual information is aligned with the visual content, acting like a relevance filter of the input text snippets. Finally, we incorporate top-$j$ sentences from NN.

\begin{figure}[t]
 \centering 
 \includegraphics[width=0.65\linewidth]{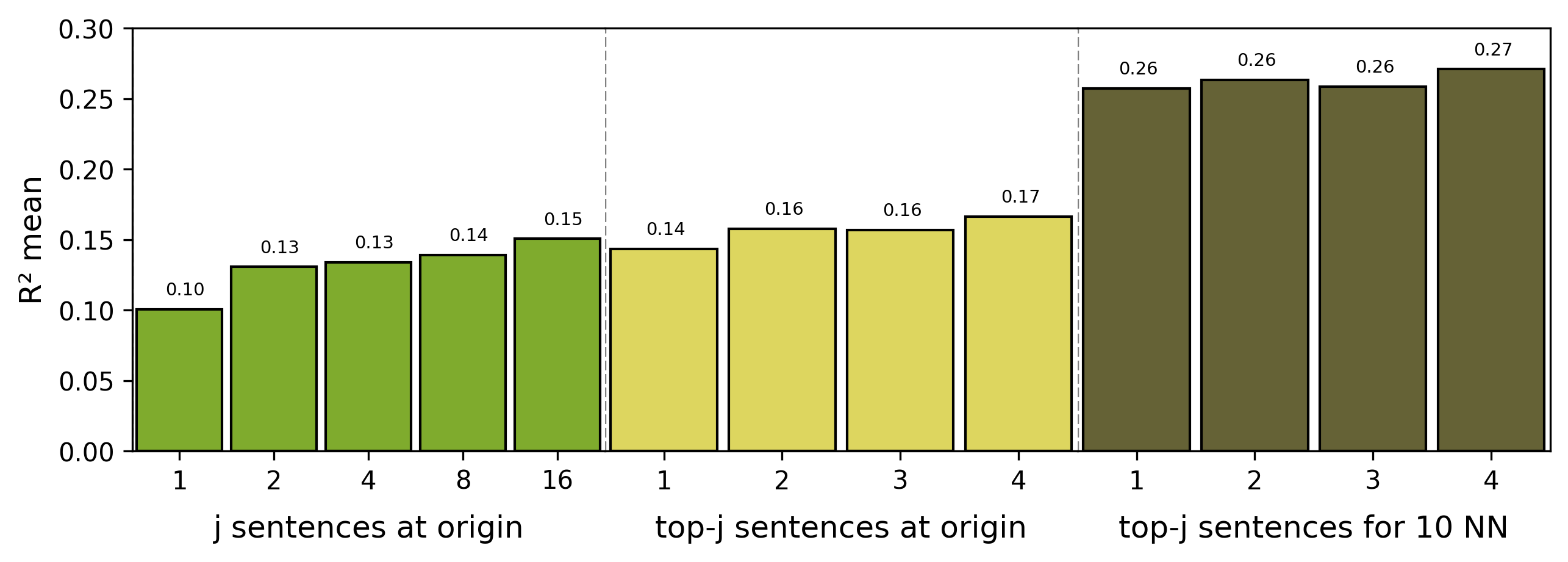}\\
 (a) Text only \\
 \includegraphics[width=0.65\linewidth]{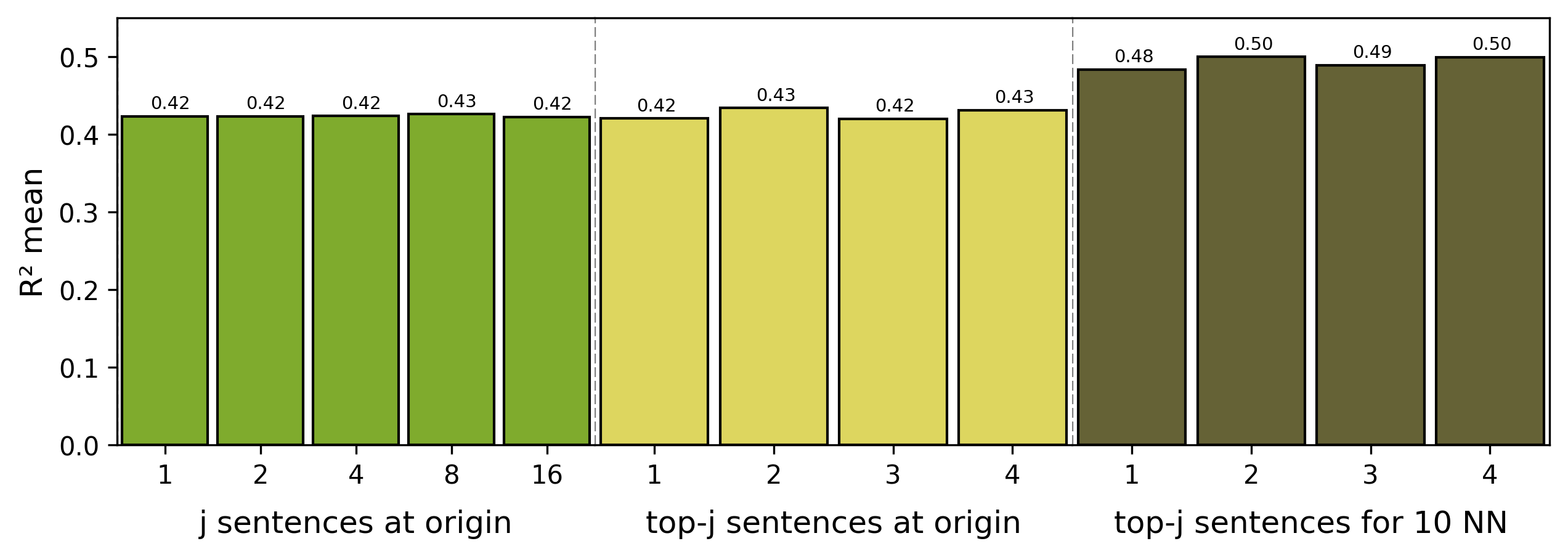}\\
 (b) Text + Images \\
 \caption{Performance for predicting SWECO25for (a) the ``Text only'' model, (b) ``Text + Images'' model, as correlation coefficient ($R^2$) for different numbers of sentences per location, and selecting top-j sentences per location based on ``Text + 1 Image'' cosine similarity with or without integrating a spatial context of $10$ NN.}
 \label{fig:text_only_ablation}
\end{figure}

Fig.~\ref{fig:text_only_ablation} (a) illustrates the results for the ``Text only'' setting. More sentences lead to an increase in the $R^2$ correlation coefficient from $0.10$ to $0.15$, as the models better capture relevant context, improving the overall results. Selecting a small number of informative sentences (top-$j$) is more effective than simply increasing the number of sentences for a given location. For instance, selecting the top-4 sentences leads to an $R^2$ value of $0.17$ compared to $0.13$ when randomly sampling 4 sentences. When incorporating top-$4$ sentences from NN, i.e. a total of 40 sentences, the performance climbs to $0.27$.
For the ``Text + Images'' model in Fig.~\ref{fig:text_only_ablation} (b), we do not see a significant increase in performance between selecting top-$j$ sentences or random sentences, probably due to the fact that the performance is driven mostly by the visual modality. However, similar to the ``Text only'' setting, integrating text from spatial neighbours leads to the overall best performance. 

\begin{figure}[t]
 \centering
 \includegraphics[width=\linewidth]{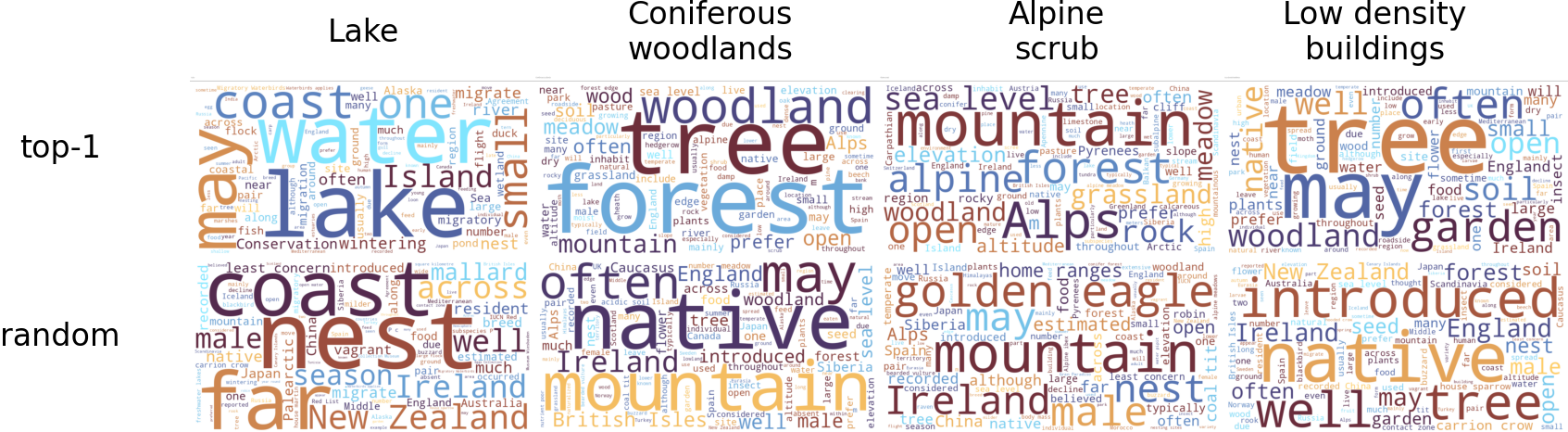} 
 \caption{Word clouds illustrating sentences associated with different EUNIS habitats from the EcoWikiRS dataset (rows) and comparison between text selection with top-j or random sentences. Colours serve only to improve readability.}
 \label{fig:wordcloud}
\end{figure}

The word clouds in Fig.~\ref{fig:wordcloud} allow us to grasp qualitatively the effects of selecting top-$j$ sentences based on similarity to the image against random sentences. The clouds are obtained with the \textit{wordcloud} Python library after filtering stop-words, i.e. commonly used words considered insignificant, such as ``the'', ``and'', and ``is'', and grouping sentences according to the ecosystem categories from EcoWikiRS. Filtering the top-1 sentences from the habitat section based on text-image similarity helps to extract objects and concepts related to land cover, and correctly describe the given EUNIS category, except for ``low density buildings'' category, where few relevant concepts are found. The relevance of the words highlighted quickly degrades when moving to random sentences, where words describing species, i.e. ``eagle'', ``nest'', ``male'', but not their living environment, occur frequently.

\subsubsection{Integrating location} \label{sec:ablation_location}

\begin{figure} [t]
 \centering
 \includegraphics[width=0.9\linewidth]{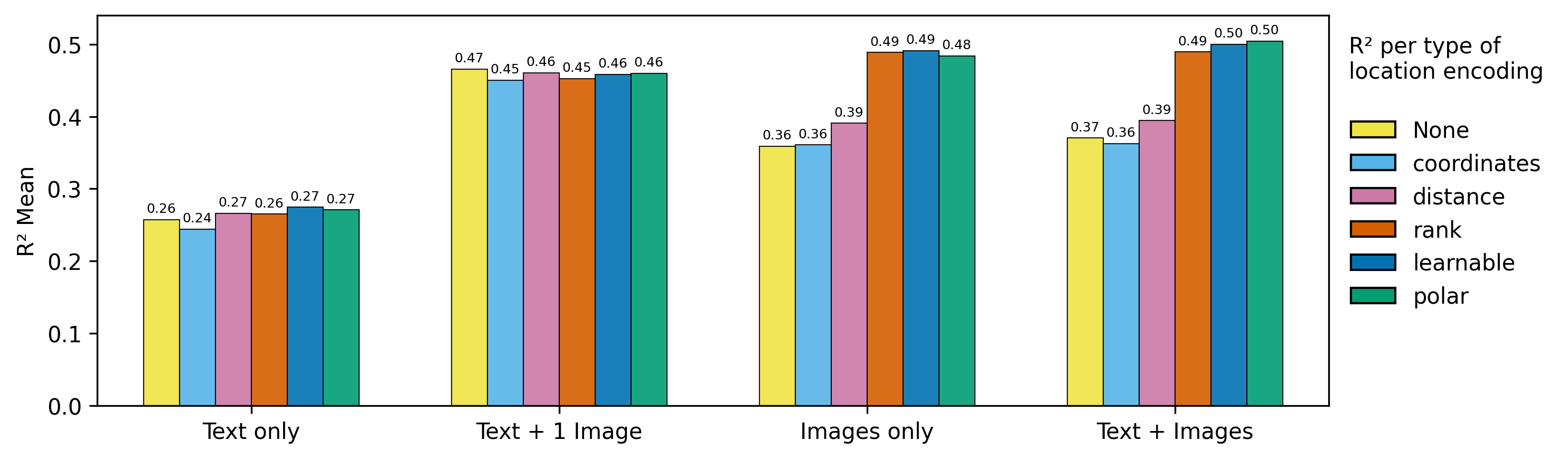}
 \caption{Comparison of different location encoding methods for each model as $R^2$ score with spatial context (k=10).}
 \label{fig:ablation_location}
\end{figure}

This section compares the performances for the different location encoding strategies described in Section~\ref{sec:method_location}. We examine whether using absolute or relative coordinates is equivalent and whether encoding the distance or the rank to the point of interest is sufficient. Results are presented in Fig.~\ref{fig:ablation_location}.

\enlargethispage{\baselineskip}
The ``Text only'' and the ``Text + 1 Image'' settings do not exhibit significant differences between the different location encoding approaches, while settings involving visual neighbours (``Text + Images'', ``Image only'') show some clear variations. One hypothesis for the absence of improvement when adding location with text is that text encodes information valid over a broad area and not specific to a location, which is potentially redundant to the location encoding, and thus does not help performance.

In the presence of visual neighbours, adding absolute coordinates does not seem to provide any advantage with respect to adding no location information at all (\textit{None}). As \textit{coordinates} encoding offers a spatial resolution of approximately $1~km$ when compressed into $512~dimensions$, it is possibly too coarse for the considered task, with the $10^{th}$ NN lying at $1~km$ on average (Suppl. Fig.~\ref{fig:similarity_vs_distance}).

We can see better performances with approaches that encode the relative position of NN to the origin location. The \textit{distance} performs only slightly better than \textit{None}, since absolute distance fails to separate neighbours located at similar ranges, but in different directions. \textit{Polar} and \textit{rank} obtain the best performance together with \textit{learnable}. \textit{Rank} and \textit{learnable} seem to be effective because they order NN by decreasing order of spatial correlation, thus theoretically of decreasing importance. \textit{Polar} encoding is more effective at encoding the spatial context than \textit{distances}: it provides roughly $40~m$ resolution for distances up to $10~km$, and the angle resolves neighbours located at similar ranges but in different directions. Overall, \textit{Polar} obtains similar or best performance, but with fewer trainable parameters, motivating our choice for other experiments.

\section{Discussion and Conclusion } \label{sec:discussion}

Geospatial data integration remains a central challenge in ecological modelling, particularly when combining gridded observations (e.g. remote sensing images) with sparse and irregularly distributed records (e.g. GBIF descriptions, camera trap images). We propose GAMMA to overcome this challenge through explicit location encoding. By representing all inputs as location-aware embeddings and leveraging a transformer encoder to model spatial and cross-modal relationships, GAMMA preserves spatial structure without the need for interpolating sparse observations into continuous spatial fields. 

We evaluate GAMMA on the simultaneous prediction of 103 environmental variables from SWECO25 \citep{kulling2024sweco25}, integrating aerial imagery and Wikipedia-derived habitat descriptions over the Swiss territory. 
Our results demonstrate that integrating two modalities, text and RS imagery, outperforms approaches relying only on one individual modality, i.e. ``Text only'' or ``Image only'' approach. Moreover, the proposed integration of spatial neighbours with geolocation embeddings consistently provides an additional gain in performance, underlining the benefits of incorporating spatial information for modelling geospatial processes. 
We analyse the benefits of each modality: Text proves to be most beneficial for variables governed by broad-scale processes (climate, edaphic, population), while variables driven by local features (vegetation) benefit mostly from features extracted from imagery. Variables that were poorly predicted from remote sensing alone (e.g., geological and hydrological variables) exhibit a marginal gain from spatial context, and no clear benefit from text integration, suggesting that insufficient relevant information was found in either modality. This points to a limitation of our experimental setup: performance is ultimately bounded by the information content of the available modalities, and some environmental processes may require dedicated observational sources not considered here.
A further limitation concerns the risk of spatial overfitting. Despite the spatial block split, the model may exploit spurious correlations or geographic sampling biases rather than learning transferable semantic content. In particular, it remains unclear whether specific “prototype” words dominate predictions at certain locations and how this affects generalisation \citep{chen2023protoclip}. Disentangling genuine semantic learning from geographic memorisation is an open problem that warrants dedicated investigation in future work.

Overall, this work demonstrates that representing heterogeneous ecological data as location-aware tokens enables flexible and scalable multimodal integration. We hope this work facilitates further integration of various data sources in ecological applications at a large scale.

\section*{Conflict of interest} 
The authors declare no conflict of interest.
\section*{Data Availability} 
\noindent The EcoWikiRS dataset is available at \href{https://doi.org/10.5281/zenodo.15236742}{https://doi.org/10.5281/zenodo.15236742}. \\
The SWECO25 variables are available at \href{https://zenodo.org/communities/sweco25/}{https://zenodo.org/communities/sweco25/}. \\

\newpage

\bibliography{_bibliography} 
\addcontentsline{toc}{section}{References}

\newpage
\section{Supplementary Material} \label{sec:appendix}

\subsection{Spatial statistics of the EcoWikiRS dataset}

\begin{figure}[h]
\centering
\includegraphics[width=0.6\linewidth]{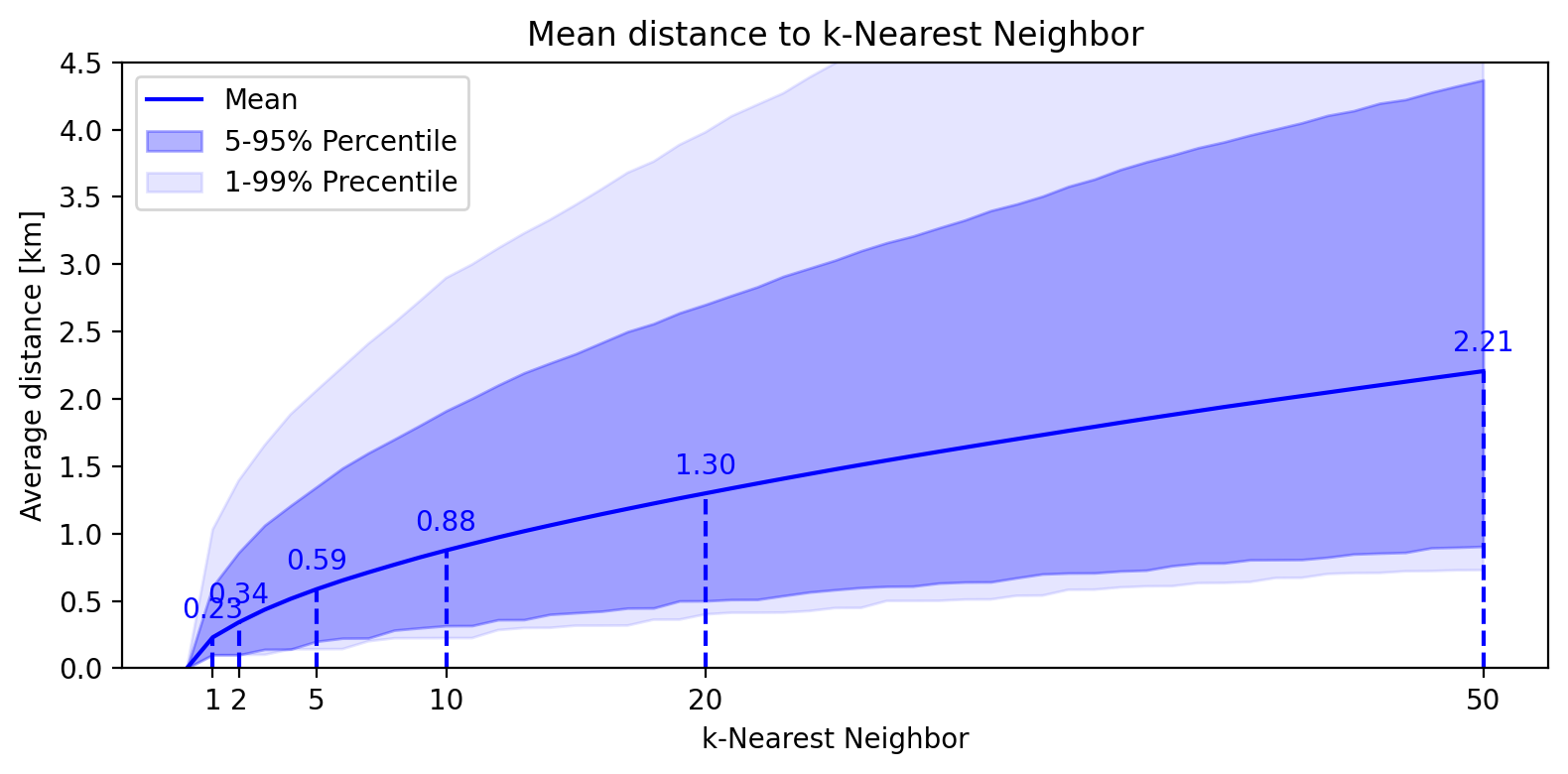}
\caption{Mean distance to $k^{th}$ nearest neighbours for samples from the EcoWikiRS dataset, with mean and distribution for $5-95$ and $1-99$ percentiles.}
\label{fig:mean_dist_to_knn2}
\centering
\includegraphics[width=0.6\linewidth]{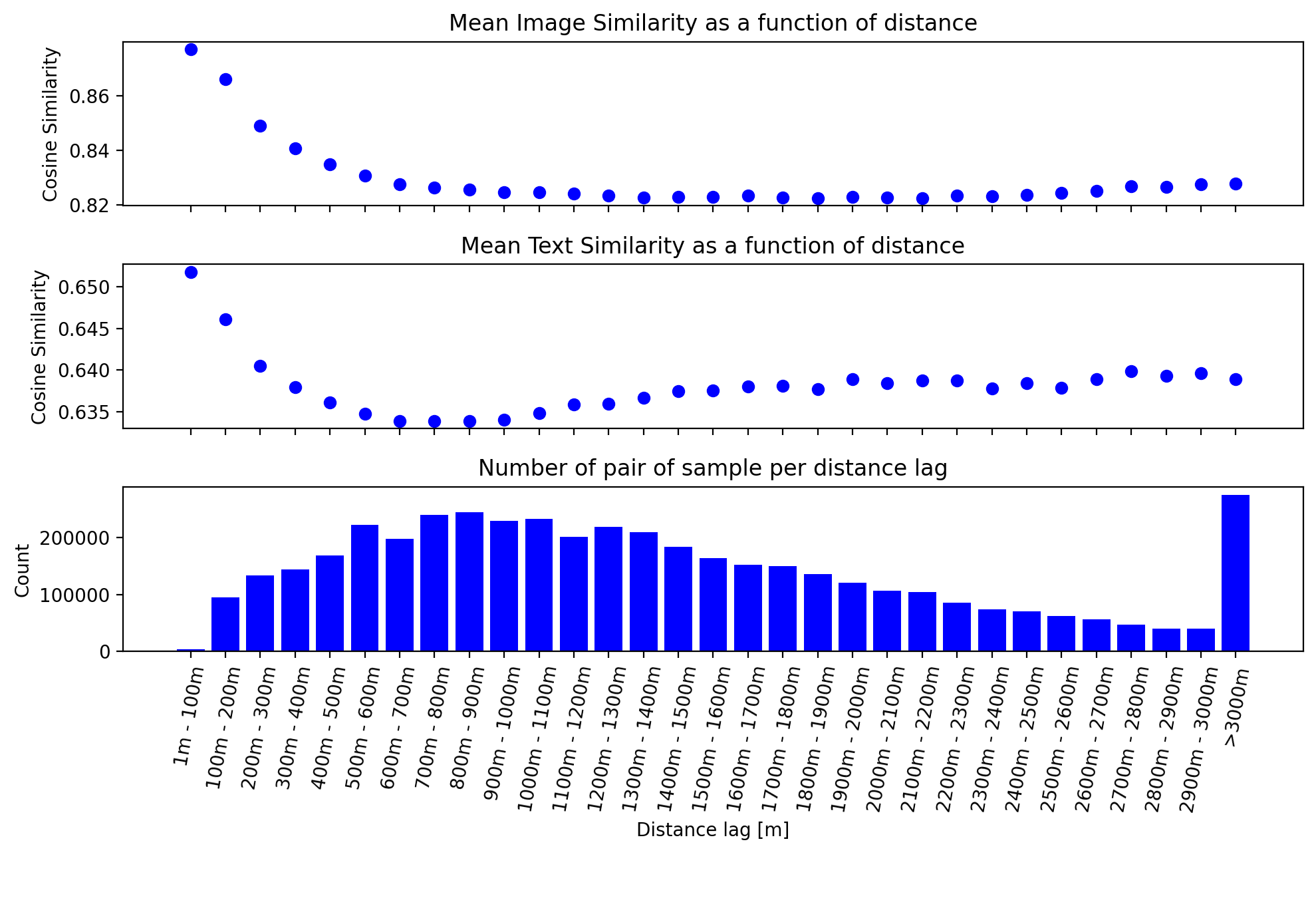}

\caption{Cosine similarity between image features (a) and text features (b) as a function of the distance between the origin and the NN samples and separated into distance lag of $100~m$. (c) Number of pairs of samples used to compute the similarity plots above. All values have been computed on the training set of the EcoWikiRS dataset. The horizontal axis is shared among the three plots.}
\label{fig:similarity_vs_distance}
\end{figure}

\noindent Fig. \ref{fig:mean_dist_to_knn2} shows the mean distance to nearest neighbours for samples from the EcoWikiRS train set. Fig. \ref{fig:similarity_vs_distance} shows the image-to-image and text-to-text cosine similarity as a function of distance for samples from the EcoWikiRS train set, encoded with the SkyCLIP model. 

\subsection{Additional results}

\subsection{SWECO25 variables selection and preprocessing}
\label{sec:sweco_docu}

\begin{tcolorbox}[colback=black!2!white,colframe=black!30!white,
 title=Box \ref{sec:sweco_docu} : SWECO layer selection,label=box:sweco,
 float=t]
 \begin{itemize}
 \item bioclim: sub period: 1981 2010.
 \item edaph: focal: 100.
 \item lulc: sub period:2023-2018 and focal: 100 or False.
 \item pop: dataset: wslhabmap and focal: 100 or False, dataset: geostat and sub period: 2020 and focal: 100 or False.
 \item geol: focal False.
 \item remote sensing: not included, since we use imagery.
 \item hydro: dataset: gwn07 and attribute: all, dataset: morph and focal: 100 or False. 
 \item trans: variable long: euclidean distance to roads and attribute: all
 \item vege: dataset Copernicus: focal: 100 or False, dataset: nfi and focal: 100 or False.
 \end{itemize}
\end{tcolorbox}

 SWECO25 is a Swiss-wide raster database at 25-meter resolution composed of $5'265$ layers. All variables are numerical, with categorical variables encoded with a distance buffer (focal) of $100m$, if not mentioned otherwise. Some SWECO variables exhibit a strong imbalanced distribution, with most values equal to zero, for example, rare land cover classes. SWECO values are extracted and averaged for each location of the EcoWikiRS dataset, following the footprint of the aerial images. The numerical values are standardised with a zero mean and a unit variance. 
 
A primary selection of $307$ variables is arbitrarily made based on the layer descriptions, spatial resolution and their interest potential for regression from EO and text data. This primary selection followed a set of rules shown in the Box \ref{box:sweco}. The final set of 103 SWECO variables is obtained by (1) removing variables with a standard deviation smaller than 0.01, (2) removing duplicate columns, defined by two columns with a correlation superior to $0.98$, (3) removing variables with low importance. This last criterion is operated by fitting a random forest model on SWECO variables to predict the EUNIS ecosystem labels from the EcoWikiRS dataset. Feature importance is computed for each SWECO variable, and a fraction of $0.5$ of variables with the lowest importance is removed, leading to the final set of 103 SWECO variables.

\input{_sweco_table}

\end{document}

%% file: _sweco_table.tex
\begin{table}[ht]
\centering
\caption{SWECO25 variables name and description (unit)}
\label{tab:metadata_sweco}
\resizebox{0.93\columnwidth}{!}{%
\begin{tabular}{ll}

SWECO25 Edaphic variables & Description (unit) \\
\hline
edaph\_eiv\_d\_25 & soil aeration (index (low to high)) \\
edaph\_eiv\_f\_25 & soil moisture (index (dry to wet)) \\
edaph\_eiv\_h\_25 & soil humus (index (humus poor to rich)) \\
edaph\_eiv\_k\_25 & continentality (index (atlantic to continental)) \\
edaph\_eiv\_l\_25 & light (index (shaded to sunny)) \\
edaph\_eiv\_n\_25 & soil nutrients (index (nutrient poor to rich)) \\
edaph\_eiv\_r\_25 & soil ph (index (acidic to alkaline)) \\
edaph\_eiv\_w\_25 & soil moisture variability (index (low to high)) \\
edaph\_modiffus\_n\_25 & nitrogen load (kilogram per hectare per year) \\
edaph\_modiffus\_p\_25 & phosphorus load (kilogram per hectare per year) \\
&\\
SWECO25 Geology variables & Description (percentage cover) \\
\hline
geol\_geotechnic\_cl1 & lakes cover  \\
geol\_geotechnic\_cl19 & massive limestones  \\
geol\_geotechnic\_cl3 & silt, clay, loess, ground moraine, surface moraine\\
geol\_geotechnic\_cl5 & gravel and sand (glacial deposit)  \\
geol\_geotechnic\_cl6 & gravel and sand (current deposit)  \\
geol\_geotechnic\_cl7 & angular gravel, blocks, slope scree  \\ 
&\\
SWECO25 Population variables. & Description (unit) \\
\hline
pop\_statpop\_2020\_populationdensity & human population density (number of inhabitants) \\
pop\_statpop\_2020\_populationdensity\_mean\_100 & human population density (number of inhabitants) \\
pop\_statpop\_2020\_populationdensity\_sum\_100 & human population density (number of inhabitants) \\
trans\_tlm3d\_dist2road\_all & euclidean distance to roads (meters) \\
&\\
SWECO25 Vegetation variables. & Description (unit) \\
\hline
vege\_copernicus\_coniferous\_100 & coniferous leaf type (percentage cover) \\
vege\_copernicus\_deciduous\_100 & deciduous leaf type (percentage cover) \\
vege\_nfi\_canopy\_max\_100 & canopy height (meters) \\
vege\_nfi\_canopy\_median\_100 & canopy height (meters) \\
vege\_nfi\_canopy\_min\_100 & canopy height (meters) \\ 
&\\ 
SWECO25 Hydrologic variables & Description (unit)\\
\hline
hydro\_gwn07\_dist2lake\_all & distance to lake (meters) \\
hydro\_gwn07\_dist2riverstrahler\_all & distance to river (meters) \\
hydro\_morph\_oekomorphology & river ecomorphology (percentage cover) \\
hydro\_morph\_oekomorphology\_cl1\_100 & river ecomorphology (percentage cover) \\
hydro\_morph\_oekomorphology\_cl2\_100 & river ecomorphology (percentage cover) \\
hydro\_morph\_oekomorphology\_cl3\_100 & river ecomorphology (percentage cover) \\ &\\
SWECO25 Bioclimatic variables & Description (unit) \\
\hline
bioclim\_chclim25\_present\_1981\_2010\_ai & aridity index (index (dry to wet)) \\
bioclim\_chclim25\_present\_1981\_2010\_bio15 & precipitation seasonality (coefficient of variation) \\
bioclim\_chclim25\_present\_1981\_2010\_bio16 & precipitation of wettest quarter (millimetre) \\
bioclim\_chclim25\_present\_1981\_2010\_bio17 & precipitation of driest quarter (millimetre) \\
bioclim\_chclim25\_present\_1981\_2010\_bio18 & precipitation of warmest quarter (millimetre) \\
bioclim\_chclim25\_present\_1981\_2010\_bio19 & precipitation of coldest quarter (millimetre) \\
bioclim\_chclim25\_present\_1981\_2010\_bio2 & mean diurnal range (degree Celsius) \\
bioclim\_chclim25\_present\_1981\_2010\_bio3 & isothermality (degree Celsius) \\
bioclim\_chclim25\_present\_1981\_2010\_bio4 & temperature seasonality (degree Celsius) \\
bioclim\_chclim25\_present\_1981\_2010\_bio7 & temperature annual range (degree Celsius) \\
bioclim\_chclim25\_present\_1981\_2010\_bio8 & mean temperature of wettest quarter (degree Celsius) \\
bioclim\_chclim25\_present\_1981\_2010\_bio9 & mean temperature of driest quarter (degree Celsius) \\
bioclim\_chclim25\_present\_1981\_2010\_etp & evapotranspiration (millimetre) \\
bioclim\_chclim25\_present\_1981\_2010\_prec & precipitation (millimetre) \\
bioclim\_chclim25\_present\_1981\_2010\_tmax & maximum temperature (degree Celsius) \\
bioclim\_chclim25\_present\_1981\_2010\_tmin & minimum temperature (degree Celsius) \\ 
\end{tabular}%
}
\end{table}

\addtocounter{table}{-1} 

\begin{table}[ht]
\centering
\caption{ continued from previous page}
\resizebox{0.9\columnwidth}{!}{%
\begin{tabular}{ll}
SWECO25 Land Use  Land Cover variables & Description (percentage cover) \\
\hline
lulc\_wslhabmap\_cl11\_100 & still water  \\
lulc\_wslhabmap\_cl12\_100 & running water  \\
lulc\_wslhabmap\_cl21\_100 & vegetated shoreline  \\
lulc\_wslhabmap\_cl22\_100 & low marshes  \\
lulc\_wslhabmap\_cl32\_100 & alluvium and moraines  \\
lulc\_wslhabmap\_cl33\_100 & scree  \\
lulc\_wslhabmap\_cl34\_100 & cliffs  \\
lulc\_wslhabmap\_cl40\_100 & grass and artificial meadows  \\
lulc\_wslhabmap\_cl41\_100 & rock slabs and limestone pavement  \\
lulc\_wslhabmap\_cl42\_100 & thermophilic dry grassland  \\
lulc\_wslhabmap\_cl43\_100 & high altitude grasslands and rough pastures  \\
lulc\_wslhabmap\_cl45\_100 & rich meadow  \\
lulc\_wslhabmap\_cl52\_100 & megaphorbs, forestry cuts  \\
lulc\_wslhabmap\_cl53\_100 & bushes and shrubs  \\
lulc\_wslhabmap\_cl54\_100 & heaths  \\
lulc\_wslhabmap\_cl60\_100 & planting  \\
lulc\_wslhabmap\_cl62\_100 & beech forest  \\
lulc\_wslhabmap\_cl63\_100 & other broadleaf forests  \\
lulc\_wslhabmap\_cl66\_100 & highland coniferous forest  \\
lulc\_wslhabmap\_cl81\_100 & woody plant crops  \\
lulc\_wslhabmap\_cl82\_100 & herbaceous crops  \\
lulc\_wslhabmap\_cl92\_100 & buildings  \\
lulc\_wslhabmap\_cl93\_100 & roads, rails, paths  \\
lulc\_wslhabmap\_cl94\_100 & parking, sport grounds  \\
lulc\_wslhabmap\_gr1\_100 & water  \\
lulc\_wslhabmap\_gr2\_100 & shorelines and wetlands  \\
lulc\_wslhabmap\_gr3\_100 & glaciers, rocks, scree and moraines  \\
lulc\_wslhabmap\_gr4\_100 & grasslands and meadows  \\
lulc\_wslhabmap\_gr5\_100 & heaths, forest edges and megaphorbs  \\
lulc\_wslhabmap\_gr6\_100 & forests  \\
lulc\_wslhabmap\_gr7\_100 & pioneer vegetation in areas disturbed by humans  \\
lulc\_wslhabmap\_gr8\_100 & plantations, fields and crops  \\
lulc\_wslhabmap\_gr9\_100 & built environments  \\ 
lulc\_geostat25\_present\_2013\_2018\_cl12\_100 & surroundings of agricultural buildings  \\
lulc\_geostat25\_present\_2013\_2018\_cl14\_100 & surroundings of unspecified buildings  \\
lulc\_geostat25\_present\_2013\_2018\_cl2\_100 & surroundings of industrial and commercial buildings  \\
lulc\_geostat25\_present\_2013\_2018\_cl31\_100 & public parks  \\
lulc\_geostat25\_present\_2013\_2018\_cl32\_100 & sports facilities  \\
lulc\_geostat25\_present\_2013\_2018\_cl39\_100 & vineyards  \\
lulc\_geostat25\_present\_2013\_2018\_cl41\_100 & arable land  \\
lulc\_geostat25\_present\_2013\_2018\_cl42\_100 & meadows  \\
lulc\_geostat25\_present\_2013\_2018\_cl43\_100 & farm pastures  \\
lulc\_geostat25\_present\_2013\_2018\_cl46\_100 & favorable alpine pastures  \\
lulc\_geostat25\_present\_2013\_2018\_cl4\_100 & surroundings of one- and two-family houses  \\
lulc\_geostat25\_present\_2013\_2018\_cl50\_100 & normal dense forest  \\
lulc\_geostat25\_present\_2013\_2018\_cl51\_100 & forest strips  \\
lulc\_geostat25\_present\_2013\_2018\_cl56\_100 & open forest (on unproductive areas)  \\
lulc\_geostat25\_present\_2013\_2018\_cl57\_100 & brush forest  \\
lulc\_geostat25\_present\_2013\_2018\_cl59\_100 & clusters of trees (on agricultural areas)  \\
lulc\_geostat25\_present\_2013\_2018\_cl62\_100 & rivers  \\
lulc\_geostat25\_present\_2013\_2018\_cl65\_100 & unproductive grass und shrubs  \\
lulc\_geostat25\_present\_2013\_2018\_cl67\_100 & wetlands  \\
lulc\_geostat25\_present\_2013\_2018\_cl69\_100 & rocks  \\
lulc\_geostat25\_present\_2013\_2018\_cl70\_100 & screes, sand  \\
lulc\_geostat25\_present\_2013\_2018\_cl8\_100 & surroundings of blocks of flats  
\\
lulc\_geostat25\_present\_2013\_2018\_cl10\_100 & surroundings of public buildings  \\ 
\end{tabular}%
}
\end{table}